\documentclass[sigconf]{acmart}
\pdfoutput=1

\AtBeginDocument{%
  \providecommand\BibTeX{{%
    \normalfont B\kern-0.5em{\scshape i\kern-0.25em b}\kern-0.8em\TeX}}}

\usepackage{graphicx}
\usepackage{subfigure}
\usepackage{amsmath}
\usepackage{algorithm}
\usepackage{algorithmic}

\usepackage{xcolor}
\usepackage{xspace} 

\DeclareMathOperator*{\argmax}{arg\,max}

\newcommand{\norm}[1]{\left\lVert#1\right\rVert}

\usepackage{hyperref}

\usepackage{amssymb}

\newcommand{\baseline}{\textsc{Ens}\xspace}
\newcommand{\ourname}{\textsc{PRISM}\xspace}
\newcommand{\ournamerand}{\textsc{PRISM}$_R$\xspace}
\newcommand{\pines}{\textsc{QO}\xspace}
\newcommand{\enspines}{\textsc{EQ}\xspace}
\newcommand{\EPPr}{\textsc{EPP$_R$}\xspace}
\newcommand{\EPQ}{\textsc{EPQ}\xspace}
\newcommand{\EPPrQ}{\textsc{EPP$_R$Q}\xspace}

\newcommand{\squishlist}{
   \begin{list}{$\bullet$}
    { \setlength{\itemsep}{0pt}      \setlength{\parsep}{3pt}
      \setlength{\topsep}{3pt}       \setlength{\partopsep}{0pt}
      \setlength{\leftmargin}{1.5em} \setlength{\labelwidth}{1em}
      \setlength{\labelsep}{0.5em} } }

\newcommand{\squishend}{
    \end{list}  }

\begin{document}

%
\title[Making targeted black-box evasion attacks effective and efficient]{Making targeted black-box evasion attacks effective and efficient}

%
\author{Mika Juuti, Buse Gul Atli, N. Asokan}
\email{{mika.juuti,buse.atli}@aalto.fi, asokan@acm.org}
\orcid{0000-0002-5093-9871}
\affiliation{%
  \institution{Aalto University, Finland}
  \streetaddress{P.O. Box 15400}
  \city{}
  \state{}
  \postcode{FI-00076}
}

%
\renewcommand{\shortauthors}{Juuti, et al.}




\begin{abstract}
We investigate how an adversary can optimally use its query budget for \emph{targeted evasion attacks} against deep neural networks in a black-box setting. 
We formalize the problem setting and systematically evaluate what benefits the adversary can gain by using substitute models. 
We show that there is an exploration-exploitation tradeoff in that \emph{query efficiency} comes at the cost of \emph{effectiveness}. 
We present two new attack strategies for using substitute models and show that they are as effective as previous ``query-only'' techniques but require significantly fewer queries, by up to three orders of magnitude. 
We also show that an \emph{agile adversary} capable of switching through different attack techniques can achieve pareto-optimal efficiency. 
We demonstrate our attack against Google Cloud Vision showing that the difficulty of black-box attacks against real-world prediction APIs is significantly easier than previously thought (requiring $\approx$500 queries instead of $\approx$20,000 as in previous works). 
\end{abstract}


%

\maketitle

\section{Introduction}
\label{sec:introduction}

The immense surge in the popularity of machine learning applications in recent years has been accompanied by concerns about their security and privacy. One such concern is \emph{evasion}. Given a machine learning classifier (\emph{victim}) and a particular input (\emph{goal}), evasion is the process of finding an \emph{adversarial example}~\cite{biggio2013evasion, szegedy2013intriguing} that is sufficiently close to the goal, but will fool the victim classifier into outputting a different class label than that for the goal. When the adversary aims for a specific misclassified label, evasion is said to be \emph{targeted}. For image classifiers, the difference between the adversarial example and goal images is imperceptible to humans.

Early techniques for finding adversarial examples against deep neural networks (DNNs) for image classification assumed a 
\emph{white-box} setting, where the adversary knows the architecture and weights of the 
victim DNN
\cite{szegedy2013intriguing, goodfellow2015explaining}. 
Since DNN cost functions are differentiable, these techniques 
calculated minimal changes (perturbations) to images which resulted in the DNN misclassifying the 
modified image. Later work addressed the
black-box setting, 
where this perturbation cannot be directly calculated. 
Papernot et al~\cite{papernot2016transferability} demonstrated that adversarial examples 
exhibit \emph{transferability}: adversarial examples for one DNN (substitute model) are likely to be adversarial to another DNN (victim model) when they are trained on datasets with a similar distribution.
Liu et al~\cite{liu2016delving} empirically demonstrated that using an ensemble of substitute models instead of a single one results in better transferability.
We call the state-of-the-art techniques in this category, such as ~\cite{liu2016delving} which rely exclusively on the use of ensembles as \baseline.
These
are \emph{efficient} in that the adversary needs to access the victim DNN only once (to test the adversarial example) but are not \emph{effective} in targeted evasion because they may not always result in successful adversarial examples.

An alternative approach for targeted black-box evasion is where the adversary repeatedly queries the prediction API of a victim DNN, and estimates its gradient solely based on the responses to the queries. We call this class of techniques~\cite{chen2017zoo, Brendel2017a, ilyas2018black}, \emph{query-only}: \pines.
%
These techniques are highly effective, reaching up to 100\%. But they are \emph{inefficient}.
For example, even state-of-the-art methods
\cite{ilyas2018prior}
require  
up to 10 000 API queries for Google's Inception v3~\cite{szegedy2016rethinking} to reach a success rate of 95.6\% assuming that the API returns probability scores for \emph{all} labels. Real-life DNN APIs often work in a \emph{partial information} (PI) setting~\cite{ilyas2018black}, where the API only returns the \emph{top-k scores} which further degrades the efficiency of \pines techniques: the state-of-the-art to the best of our knowledge is~\cite{ilyas2018black} which reports requiring 350,000 queries reach a success rate of 93.6\% on this network in a PI setting.
\begin{figure}%
\includegraphics[width=1.\columnwidth]{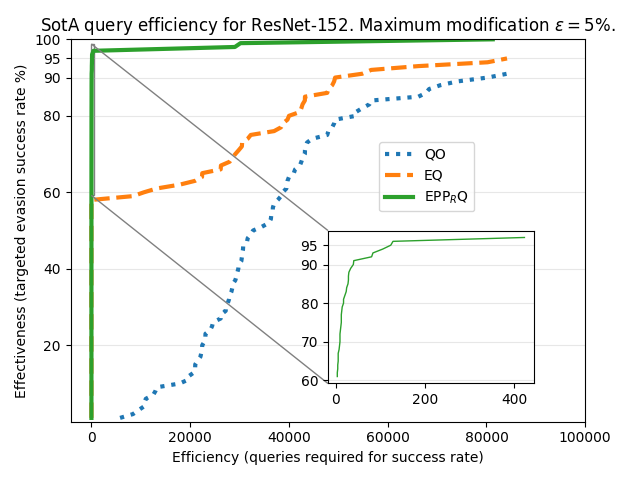}
\caption{
Comparison of targeted black-box attacks against a partial information ResNet-152 model using different strategies: \pines~\cite{ilyas2018black} vs. \enspines (using MIFGSM for gradient estimation~\cite{dong2018boosting}, followed by \pines), vs. a fully agile adversary \EPPrQ switching through all methods 
  (Sec.\ref{sec:strategy}).
}
\label{fig:2figsB}%
\end{figure}%

In this paper, we ask if we can design targeted black-box evasion techniques that are simultaneously efficient and effective. We argue that a realistic solution should (a) be designed for the partial-information setting since real-life APIs often use this setting\footnote{e.g. \url{https://cloud.google.com/vision/}, \url{https://clarifai.com/demo}}; and (b) use ensembles because of the widespread availability of \emph{public, pre-trained} models.
Our contributions are as follows:
\squishlist
\item We show that \baseline followed by \pines (\enspines) can outperform pure \pines techniques (Sec.~\ref{sec:strategy}).
\item We present \ourname and \ournamerand, two new targeted black-box evasion techniques that (a) starts from an input with the target label, and (b) repeatedly query the victim API (within a query budget) while concurrently using an ensemble to estimate victim's gradient in a PI setting. (Sec.~\ref{sec:partial}). We show that the effectiveness of \ourname is comparable to \pines on publicly available ImageNet models while typically requiring \emph{fewer victim queries by three orders of magnitude} (Sec.~\ref{sec:efficiency}).
\item We systematically compare the different evasion approaches to show that an \emph{agile adversary} can switch through these approaches in a particular order to achieve optimal efficiency, e.g. 13219 queries on average to reach the effectiveness 94\% on Inception v3 in PI setting (Sec.~\ref{sec:strategy}).
  
\item We demonstrate the real-world applicability of \ourname by producing a targeted adversarial example against Google Cloud Vision, reducing the
number of queries required from an estimate of 20,000 queries~\cite{ilyas2018black}
to approximately 500 (Sec.~\ref{sec:real}). 
\squishend
Figure~\ref{fig:2figsB} shows the benefit of attacker agility by comparing different strategies: \pines vs. \enspines vs. a fully agile adversary \EPPrQ who switches through all methods (\baseline, \ourname, \ournamerand and \pines). In Section~\ref{sec:discussion}, we discuss possible reasons why the \ourname approach is effective.

\section{Preliminaries}
\label{sec:preliminaries}

We first lay out definitions of frequently used concepts and functions in this work,
with the focus on the image domain. 

\subsection{API definitions}
\label{sec:definitions}

\paragraph{{Classifier},}
\texttt{clf}: a function 
that maps an arbitrary input \emph{image}
$\mathbf{x} \in {\rm [0,1]}^{{\rm c} \times {\rm w} \times {\rm h}}$ 
to a vector of probabilities 
$\mathbf{p} \in [0,1]^{{\rm N}}$,
denoting \texttt{clf}'s confidence in 
assigning $\mathbf{x}$ to any of the pre-defined classes $\{{\rm 1}, \dots, {\rm N}\}$. The elements of $\mathbf{p}$ sum to 1. 
Here ${\rm c}$ refers to the number 
of color channels,
${\rm w}$ to the width and ${\rm h}$ to the height
of input $\mathbf{x}$. 
A deep neural network \texttt{dnn} is a particular type of  \texttt{clf} parameterized with $M$ sequential functions:
\begin{equation}
\texttt{dnn} (\mathbf{x}) = f_M \circ f_{M-1} \circ \dots f_{2} \circ f_1 \circ \mathbf{x} ,
\end{equation}

where each function $f_i (\mathbf{z})$ can be expressed as $\sigma_i (\mathbf{w}_i^{T} \mathbf{z} + \mathbf{b}_i)$,
where $\sigma_i$ is a (nonlinear) function, $\mathbf{w}_i$ is a weight matrix and $\mathbf{b}_i$ is a bias vector.
In this work, we focus on \texttt{dnn}-based image classifiers. In a typical 
\texttt{dnn}, $\sigma_i$ is selected as a differentiable function. Therefore, it can calculate the gradient of classification error ($\nabla_x \texttt{dnn}$)
in order to minimize this error easily in the training 
procedure~\cite{goodfellow2016deep}.  

\paragraph{{Preprocessor},}
\texttt{pre}: A function that receives an input $\mathbf{x} \in {\rm I\!R}^{{\rm c} \times {\rm h} \times {\rm w}}$ 
and produces an output $\mathbf{x}' \in {\rm I\!R}^{{\rm c'} \times {\rm h'} \times {\rm w'}}$, and is primarily used for formatting, normalizing and resizing input $\mathbf{x}$ before it is classified by \texttt{dnn}, since \texttt{dnn} only processes fixed input sizes. 
\texttt{pre} can be used to break the differentiability property of \texttt{dnn}
 and act as a form of defense (shattered gradients, \cite{athalye2018obfuscated}). 

\paragraph{{Postprocessor},}
\texttt{post}: A function that receives input $\mathbf{p} \in {\rm I\!R}^{{\rm N}}$ and
produces an output $\mathbf{p'} \in {\rm I\!R}^{{\rm N'}}$. It is used for 
formatting the output $\mathbf{p}$ of \texttt{dnn}, both for readability and
limiting information from \texttt{dnn}. Common choices are:  
\begin{itemize}
\item Identity function: $\texttt{post}_{I} (\mathbf{x}) \leftarrow \texttt{I}(\mathbf{x}) ~\forall x$
\item Label-only: $\texttt{post}_\mathcal{L} (\mathbf{x}) \leftarrow \argmax (\mathbf{x})$, i.e. the index $i \in \{1, ..., N \}$ with the largest value in $x$.
\item Top-k: $\texttt{post}_{\rm k} (\mathbf{x}) \leftarrow \mathbf{w}_{\rm k} \cdot \mathbf{x} \in {\rm I\!R}^{{\rm N}}$, such that $\mathbf{w}_{\rm k, i} \leftarrow 1$ iff $i$ is among the ${\rm k}$ first values in \texttt{sorted}($\mathbf{x}$), where $\mathbf{x}$ is sorted in descending order. $\mathbf{w}_{\rm k, i} \leftarrow 0$ for others.
\end{itemize}

\paragraph{{API}:}
\label{sec:api}
A combination of 
\texttt{pre}, 
\texttt{dnn} and 
\texttt{post}
that responds to arbitrary client, \textsl{CLI}, queries $\mathbf{x}$ with the following response:
\begin{equation}
{} _ \texttt{pre} \texttt{API} _ \texttt{post} (\mathbf{x}) = \texttt{post} \circ \texttt{dnn} \circ \texttt{pre} \circ \mathbf{x}
\label{label:api}
\end{equation}
An ideal \texttt{API} always responds to \textsl{CLI}'s queries $\mathbf{x}$, as long as $\mathbf{x}$ is not malformatted.
Prior work in black-box adversarial examples however typically
do not use preprocessing on APIs  
$\texttt{pre}(\mathbf{x}) = \texttt{I}(\mathbf{x}) = \mathbf{x}$.
In this paper, we only focus on preprocessing that resizes input correctly for 
\texttt{dnn}. 


\paragraph{{White-box access}:}
\textsl{CLI} knows the precise definition of every intermediate function applied on any arbitrary input $\mathbf{x}$. 
Moreover, $\texttt{pre}(\mathbf{x}) = \texttt{post}(\mathbf{x}) = \texttt{I}(\mathbf{x})$,
i.e. \textsl{CLI} has access to all of \texttt{dnn}'s outputs. 

\paragraph{{Gray-box access}:}
\textsl{CLI} does not know the full definition of
$\texttt{pre}$ and $\texttt{post}$ or the network parameters of \texttt{dnn}, 
but may know other information such as the architecture, hyper-parameters, training method and the training set of \texttt{dnn}~\cite{meng2017magnet}. 


\paragraph{{Black-box access}:}
\label{sec:black-box}
\textsl{CLI} 
does not know the exact forms of any intermediate functions. 
Different authors define the minutiae of black-box \texttt{API} differently, 
we adapt these as follows (in order of decreasing privilege):
\begin{itemize}
\item \textsl{Maximum information }($\texttt{API}_{I}$):
\texttt{dnn} is secret, 
while \textsl{CLI} has access to probabilities or logits from \texttt{dnn} for arbitrary input $\mathbf{x}$ ~\cite{ilyas2018prior}. 
\item \textsl{Partial information }($\texttt{API}_{k}$):
\textsl{CLI} has access to top-{\rm k} output from \texttt{dnn} for arbitrary input $\mathbf{x}$~\cite{ilyas2018black}. 
Generally, a realistic \texttt{API} has a long probability list 
$\mathbf{p} \in {\rm I\!R}^{{\rm N}}$ with $N \geq 1000$ and returns a small subset of this list ($k \ll N$).
\item \textsl{Label-only }($\texttt{API} _ {\mathcal{L}}$): 
\textsl{CLI} has access only to labels from \texttt{dnn} for arbitrary input $\mathbf{x}$~\cite{ilyas2018black}. 
\end{itemize}

\subsection{Adversarial example definitions}
\label{sec:adv}
\paragraph{Adversarial example:} The adversary aims to produce an adversarial 
example $x_{\rm adv}$ that is very similar to a goal image $x_{\rm goal}$, but \emph{evades}
classification by \texttt{API}: $\texttt{API}(x_{\rm adv}) \neq \texttt{API} (x_{\rm goal})$ (non-targeted evasion). 
The similarity between $x_{\rm goal}$ and $x_{\rm adv}$ is often evaluated by an $L_p$-norm~\cite{sharif2018suitability}: $ \norm{x_{\rm goal} - x_{\rm adv}}_{\rm p}$. 
In this work, we set $p=\infty$ as is common. In targeted evasion, adversarial examples require that 
$ y' \leftarrow \texttt{API}(x_{\rm adv})$
for a pre-defined class $y' \neq \texttt{API} (x_{\rm goal})$. Targeted evasion is described in Equation~\ref{eq:adv}.

\begin{equation}
\label{eq:adv}
\begin{split}
y' &\leftarrow \texttt{API} (x_{\rm adv})\\
\text{s.t.~} & \norm{x_{\rm goal} - x_{\rm adv}}_\infty \leq \epsilon \\
 x_{\rm goal}, x_{\rm adv} \in &~[0,1]^{\rm{c} \times w \times h},
\end{split}
\end{equation}
where $\epsilon$ is the allowed \emph{perturbation} size.

\textsl{White-box attacker}, $\mathcal{A}_{\rm Wbox}$, is a malicious client 
that has white-box access to \texttt{API} which it tries to evade. Since $\mathcal{A}_{\rm Wbox}$ knows the precise definition of every intermediate function in \texttt{dnn} inside \texttt{API}, it is able to calculate the gradient  of the classification error with respect to the input image: $\nabla_x \texttt{dnn}$.   
It uses this information to modify $x_{\rm adv}$.
Existing evasion methods such as single-step fast gradient sign method FGSM~\cite{goodfellow2015expressing} and iterative version of it 
I-FGSM~\cite{kurakin2016adversarial} 
find adversarial example $x_{\rm adv}$ 
by maximizing the cross entropy loss function 
of \texttt{dnn} under $L_\infty$-norm.

\textsl{Black-box attacker}, $\mathcal{A}_{\rm Bbox}$, is a malicious client that has black-box access to $\texttt{API}'$ that it tries to evade. 
Rotations and translation operations are often enough to create non-targeted evasion
 in black-box APIs~\cite{engstrom2018rotation}. 
 However, in order to create targeted adversarial examples with a small perturbation,
 an approximation to the gradient
information, $\hat{\texttt{G}} _ \texttt{dnn}'$, of the target classifier $\texttt{dnn}'$ becomes necessary. 
There are two predominant ways of obtaining this information: (a) gradient 
approximation through \textsl{transferability} and (b) \textsl{finite-difference methods}. 


\paragraph{Transferability:} 
An adversarial example developed for evading one \texttt{API} (\texttt{dnn}) can be also adversarial to another  $\texttt{API}'$  ($\texttt{dnn}'$), i.e.

\begin{equation}
\label{eq:adv_trans}
\begin{split}
y' &\leftarrow \ \texttt{API}' (x_{\rm adv})\\
\text{s.t.~} & \text{Equation (\ref{eq:adv})}
\end{split}
\end{equation}

Recently, Adam et al.~\cite{adam2019reducing} found that the cosine similarity 
between the gradient $\nabla_x \texttt{dnn'}$ 
 and the available gradient $\nabla_x \texttt{dnn}$ is a reliable estimator for transferability. 
Thus, implicitly the following approximation occurs during transferability: 
\begin{equation}
\nabla_x \texttt{dnn} \approx \nabla_x \texttt{dnn}'
\label{eq:nabla}
\end{equation}
Liu et al.~\cite{liu2016delving} states that transferability depends on the architectural similarity of $\texttt{dnn}'$ and \texttt{dnn}. 

\begin{figure}[tbp]
\centering
\includegraphics[width=.25\textwidth]{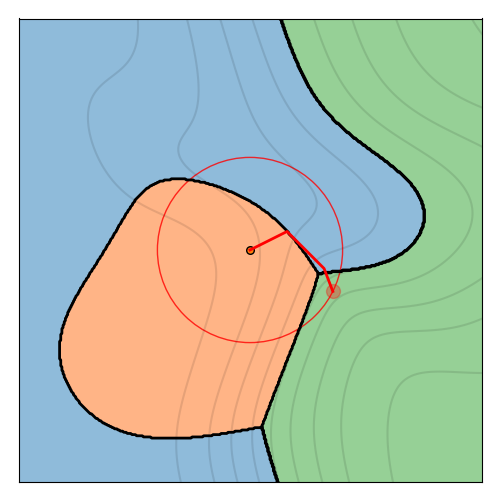}
\caption{Decision boundaries of an MLP classifier (3 classes). 
  Targeted evasion with \baseline, starting from \emph{goal} at origin. 
Maximum modification $\epsilon$ indicated with a red square.
Toy example: perturbations bounded with an $L_2$-norm.
}
\label{fig:schema-0}
\end{figure}

\paragraph{Baseline transferability attack:} Adversary  
$\mathcal{A}_{Bbox}$ has a label-only black-box access to 
$\texttt{API}'$. It tries to evade 
$\texttt{API}'$ by creating adversarial examples using many available \texttt{API}s
 and relies on the transferability property holds for the attacker's adversarial examples. 
Current state-of-the-art transferability attacks use ensembles of pre-trained DNNs  
 as \texttt{dnn}~\cite{liu2016delving}. 
A momentum-iterative version of FGSM (MIFGSM)~\cite{dong2018boosting}, won both the targeted and non-targeted evasion  
 competition at the NIPS workshop in 2017 and is since then considered to be
 the strongest $L_\infty$-bounded transferability attack.
\emph{ We call this evasion method \baseline. }
We provide an illustration for \baseline evasion for a toy example 
in Figure~\ref{fig:schema-0}. 
The classifier in the figure is a multilayer perceptron (MLP)~\cite{murphy2012machine} with three classes. 

The effectiveness of transferability attacks is usually reported in terms of whether the victim model is fooled when it is queried with the adversarial example \emph{once}. 
In Section~\ref{sec:effectiveness}, we explore whether allowing multiple queries 
brings any benefit to the adversary. 
\baseline starts querying $\texttt{API}'$ once it has reached the maximum allowed
modification $\epsilon$, as an effort to reduce the number of queries. 



\paragraph{Finite-difference methods,} 
FDM, which are also known as zero-order optimization methods, directly estimate gradients 
$\hat{\texttt{G}} _ {\texttt{dnn}'}(x)$ for a target 
 $\texttt{API}'$ by making repeated queries around $x$ \cite{chen2017zoo, ilyas2018black, tu2018autozoom, ilyas2018prior} and recording minute differences in the returned values. 
The baseline assumption is that $\texttt{API}'$ returns maximum information ($\texttt{API}'_{I}$). However, this is not a realistic assumption in practice. For example, Google Vision and Clarifai both return only top-k results from \texttt{dnn}. 

Some papers~\cite{Brendel2017a, ilyas2018black} report results under these more realistic APIs. 
The efficiency of FDMs under these API models degrades, e.g. 
\cite{ilyas2018black} evaluates that the median number of queries grows to approximately 
50,000 per example with 
$\texttt{API}'_{\rm k=1}$, 
whereas they were found to be approximately 10,000 per example with $\texttt{API}'_{I}$.
The number of queries further grows to 2.7 million with $\texttt{API}'_{\mathcal{L}}$

Evasion against $\texttt{API}'_{\rm k=1}$ 
requires a change in
the creation of targeted adversarial examples. Since 
$\texttt{API}'_ {\rm k=1}$ 
does not return feedback on any other class than the top-1 label, $x_{\rm adv}$ must always
remain as the top-1 label. 
Evasion needs
to initialize the adversarial image with another image $x_{\rm start}$ of the target class $y'$: 
\begin{equation}
\begin{split}
x_{\rm adv}^0 &\leftarrow x_{\rm start} \\
\text{s.t.~} &\texttt{API}'_ {\rm k=1}(x_{\rm adv}^0) = y',
\label{eq:pi-start}
\end{split}
\end{equation}
We define j-th iteration of an  adversarial image $x_{\rm adv}^j$ as a series of $j$ modifications
of $x_{\rm adv}^0$ towards the original image $x_{\rm goal}$.
The distance between j-th adversarial image  $x_{\rm adv}^j$ and $x_{\rm goal}$ gradually decreases when $j$ increases,
so that the evasion process eventually ends with $x_{\rm adv}^i$ ($i \geq j$) that is within a $\epsilon$-distance
from a goal image $x_{\rm goal}$:

\begin{equation}
\begin{split}
y' &\leftarrow \texttt{API}'_ {\rm k=1}(x_{\rm adv}^i) \\
\text{s.t.~} &\norm{x_{\rm adv}^i - x_{\rm goal}}_\infty \leq \epsilon\\
& 1 \leq i \leq B,
\label{eq:pi-end}
\end{split}
\end{equation}
where $B$ is maximum number of queries allowed by the $\texttt{API}'_ {\rm k=1}$. 
\cite{ilyas2018black} reported success at attacking Google Cloud Vision (GCV, December 2017) 
using this strategy, 
successfully fooling the system with perturbation size $\epsilon = 25.5 / 255$.
However, success came at a high cost: approximately some 170 gradient estimation steps, 
which we estimate is approximately 20,000 queries for one sample\footnote{URL: \url{https://www.labsix.org/media/2017/12/20/skier_adv.png}. GCV has been re-trained since then and the sample they provide does not evade GCV in March, 2019.} they provide. 
\emph{We call this evasion method \pines. }
\pines relies on the black-box optimization technique Natural Evolution
Strategies (NES) by Wiestra et al.~\cite{wierstra2008natural}, which is used for gradient estimation. 
This is a common limitation in all \pines methods, since they rely on querying
$\texttt{API}'$ at all steps.  
The authors report a success rate of $93.6\%$ on creating targeted adversarial examples for
Inception v3~\cite{szegedy2016rethinking}, using a budget of 1 million queries.

\begin{figure}[tbph]
\centering
\includegraphics[width=.25\textwidth]{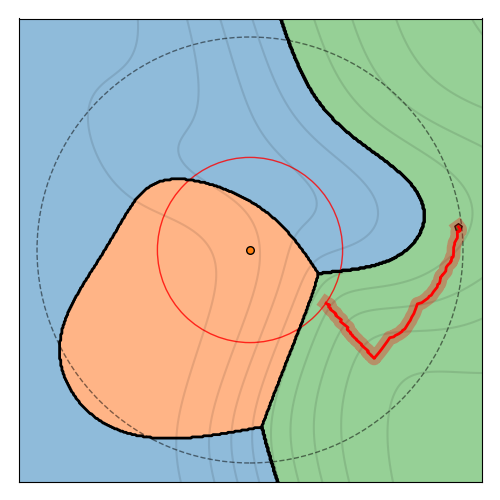}
\caption{Decision boundaries of an MLP classifier (3 classes). 
  Targeted evasion with \pines, starting from \emph{start} and gradually
approaching \emph{goal} at origin. 
Maximum modification $\epsilon$ indicated with a red square.
Toy example: perturbations bounded with an $L_2$-norm.
}
\label{fig:schema-2}
\end{figure}


\section{\ourname}
\label{sec:approach}
We further extend upon the previously mentioned methods, but evaluate a more realistic
 adversary. Our motivation is as follows. 
Due to the recent trend of striving for reproducibility in machine learning, 
tens or hundreds of pretrained models are available to the adversary.
At the same time, model owners limit the APIs to only reveal partial information (Sec.~\ref{sec:black-box}). 
We focus on an adversary models where:
For that we propose the \ourname method:
a novel way of attacking such APIs. 
We then define our evaluation criteria: \emph{success} and \emph{pareto-efficiency}. 
We begin with defining the adversary next. 


\subsection{Adversary model}
\label{sec:adv-model}

\paragraph{Goal and capabilities:}
The goal of the adversary $\mathcal{A}_{\rm Bbox}$ is to evade a black-box $\texttt{API}'$ hosting $\texttt{dnn}'$ which classifies ImageNet images. 
$\mathcal{A}_{\rm Bbox}$ can query the $\texttt{API}'$ multiple times and continue attacking 
until a successful evasion attack
is encountered or a reasonable query budget $B$ is exceeded. 
$\mathcal{A}_{\rm Bbox}$ 
has access to several 
other publicly available pre-trained ImageNet DNNs~\cite{SqueezeNet, he2016deep, huang2017densely, Simonyan14c, szegedy2016rethinking}
that it is free to combine in any way to reach its goal. 
 $\mathcal{A}_{\rm Bbox}$ is \emph{agile}, meaning that  
given a set of evasion methods, $\mathcal{A}_{\rm Bbox}$  will choose the method that is 
most likely to result in evasion with a minimum number of query. 
Given a setting $\texttt{s} = (x_{\rm start}, x_{\rm goal}, y', \texttt{API}')$, $\mathcal{A}_{\rm Bbox}$ chooses an evasion method $m_i$ while considering the query budget $B$, and produces targeted adversarial example $x_{\rm adv}$.

\paragraph{Attack surface:}
$\mathcal{A}_{\rm Bbox}$ attacks a partial information $\texttt{API}'_ {\rm k=1}$: 
it has the format as defined in Equation~\ref{label:api}
 and returns the top-1 output from $\texttt{dnn}'$. 
 $\mathcal{A}_{\rm Bbox}$ does not know the native image size of
$\texttt{dnn}'$ nor the resizing operator in $\texttt{pre}$. 
$\texttt{API}'$ 
is an ideal API: it always returns responses to queries. 

\subsection{\ourname attack technique}
\label{sec:partial}
\begin{figure}[tbph]
    \centering
    \includegraphics[width=0.25\textwidth]{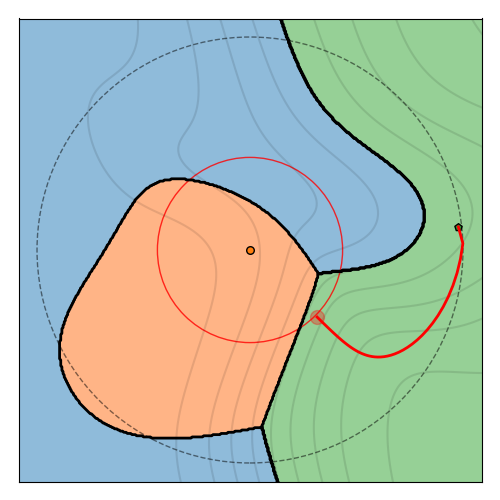}
\caption{Decision boundaries of an MLP classifier (3 classes). 
Targeted evasion with \ourname, starting from \emph{start} and gradually
approaching \emph{goal} at origin. 
Maximum modification $\epsilon$ indicated with a red square.
Toy example: perturbations bounded with an $L_2$-norm. 
}    
\label{fig:schema-1}
\end{figure}

Next, we describe \ourname (\emph{p}a\emph{r}tial \emph{i}nformation \emph{s}ubstitute \emph{m}odel), an approach for targeted evasion technique that combines strengths we 
identified in \baseline and \pines. 
We start by providing an illustration of \ourname in Figure~\ref{fig:schema-1}. 
Conceptually, \ourname is similar to \pines in Figure~\ref{fig:schema-2}: 
it starts the evasion process with image $x_{\rm start}$ of the target class $y'$ (Equation~\ref{eq:pi-start}) and finishes when it finds a solution that is within an $\epsilon$-distance to \emph{goal} image $x_{goal}$ (Equation~\ref{eq:pi-end}).
Initially, $x_{\rm start}$ is at distance $\norm{x_{\rm start} - x_{\rm goal}}_\infty = d \gg \epsilon$
The method consists of
several iterations of increasing the classification likelihood of the 
j-th iteration $x_{\rm adv}^j$ 
by stepping in the direction of the 
approximated gradient $\hat{\texttt{G}} _ {\texttt{dnn}'}$ 
and then projecting it closer to $x_{\rm goal}$. 
The procedure continues until the distance between $x_{\rm adv}^j$ and $x$ 
decreases to $\epsilon$, and $x_{\rm adv}$ is classified as the
target class $y'$ by $\texttt{API}'_ {\rm k=1}$ or until a query budget $B$ has been exceeded. 
Although the process of finding $x_{\rm adv}$ is similar to 
\pines~\cite{ilyas2018black}, the gradient estimator comes via substitute model  
ensembles and MIFGSM as is done in \baseline~\cite{dong2018boosting}. 
We detail pseudocode for \ourname in Algorithm~\ref{alg:pi-algorithm}. 

The following steps highlight the differences between \ourname and \pines algorithms:
\begin{enumerate}
\item[S$1.$] 
We set  $\hat{\texttt{G}} _ {\texttt{dnn}'} (x) \leftarrow \text{ \baseline}(x)$ in \ourname, which comes from 
the ensemble models and therefore does not consume queries. With default settings, 
\pines uses 100 queries to determine $\hat{\texttt{G}} _ {\texttt{dnn}'}  (x)$
via finite-difference methods.
\item[S$2.$] \pines uses line search~\cite{murphy2012machine} to find an
 appropriate step size with the purpose of reducing queries.
\ourname does not do this, since no queries are used. 
\item[S$3.$] Since \ourname relies on gradient estimates from substitute models,
we can aggressively avoid using queries until $d$ has reached $\epsilon$. \pines 
cannot do this, as it needs to remain inside the top-k region in order to calculate
gradients. 
\end{enumerate}

\begin{algorithm}
    \caption{\ourname attack technique}
    \label{alg:pi-algorithm}
    \begin{algorithmic}
        \REQUIRE Target $\texttt{API}_{\rm k}(x')$, 
        goal image $x_{\rm goal}$, 
        targeted class $y'$, 
        starting image of target class $x_{\rm start}$, 
        gradient estimator $\hat{\texttt{G}}(x')$, 
        goal $L_\infty$-distance $\epsilon \leftarrow 0.05$,
        patience $C \leftarrow 5$,
        update threshold $t_{adv} \leftarrow 20\%$,
        initial $L_\infty$-distance limit $d \leftarrow 0.50$,
        $\delta_\epsilon \leftarrow 0.005$.
        \STATE [ ]         
        \STATE [ ] $x' \leftarrow \textrm{Clip}(x', x_{\rm start}-d, x_{\rm start}+d)$ 
        \WHILE{$d > \epsilon$ and $\texttt{API}_\mathcal{L}(x_{\rm adv}) \neq y'$}
            \STATE [S$1.$] $g' \leftarrow \hat{\texttt{G}}(x')$ \COMMENT{gradient estimation} 
            \STATE [S$2.$] $x' \leftarrow x_{\rm adv} + \eta \cdot sign ( g' )$.
            \STATE [    ] $x' \leftarrow \textrm{Clip}(x', x_{\rm goal}-d, x_{\rm goal}+d)$.                                
            \IF{$d = \epsilon$}               
            \STATE topk $\leftarrow \texttt{API}_{\rm k}(x')$.
            \ELSE
            \STATE [S$3.$] topk $\leftarrow [y']$. \COMMENT{Pseudolabel until $d \gets \epsilon$}            
            \ENDIF
            \IF{$y' \in$ topk}   
                \STATE [  ] $x_{\rm adv} \leftarrow x'$
                \IF{ topk$[y']$ $ \ge t_{adv}$} 
                    \STATE [  ] $x_{\rm backtrack} \leftarrow x_{\rm adv}$ \COMMENT{update high-confidence $x'$}
                \ENDIF
                \STATE $d \leftarrow max(\epsilon,\epsilon-\delta_\epsilon)$
            \ELSE
                \STATE [  ] $c \leftarrow c+1$
                \IF{$c > C$ }
                    \STATE [  ] {$x_{\rm adv} \leftarrow x_{\rm backtrack}$}
                \ENDIF
            \ENDIF 
        \ENDWHILE
    \end{algorithmic}
\end{algorithm}

It is important to note that \ourname succeeds only as long as Relation~\ref{eq:nabla} holds. 
For this, we calculate $\hat{\texttt{G}} _ {\texttt{dnn}'} $ using a large ensemble (approximately 10 models). 
We further define \ournamerand as a variant of \ourname,
with an emphasis on diversity. Instead of always using the same ensemble for 
gradient estimation like in \ourname, \ournamerand
subsamples a random number of ensembles and calculates the gradient with these ensembles using \baseline method. 


\section{Evaluation}
\label{sec:evaluation}

\subsection{Experimental setup}

We first describe our experimental setup, target models and black-box evasion methods.
We take the 100 examples from ImageNet as our initial images. These images were used
in prior research~\cite{liu2016delving}\footnote{\label{foot1}\url{https://github.com/sunblaze-ucb/transferability-advdnn-pub/blob/master/data/image_label_target.csv}}, where they were chosen 
randomly from the ImageNet validation set, 
such that they were classified correctly by all models in their experiments.
We use these images as $x_{\rm goal}$. 
Our adversary model also specifies $x_{\rm start}$. Since the dataset~\cite{liu2016delving} 
 does not consider the partial information setting, we choose $x_{\rm goal}$,
$x_{\rm start}$ and target class $y'$ from this dataset according to Algorithm~\ref{alg:experiment}.
We use this setup for all 100 experiments. 
%

\begin{algorithm}
    \caption{Experiment Setup}
    \label{alg:experiment}
    \begin{algorithmic}
        \REQUIRE{dataset $\mathcal{D}$ with 100 entries $e_{\rm i}$, each entry in format $e_{\rm i} = (x_{\rm i},c_{\rm i})$,
        where $x_{\rm i}$ refers to an image, and $c_{\rm i}$ is the assigned class}
        \FOR{$i \gets 0$ to $100$}
        \STATE{$x_{\rm goal} \gets x_{\rm i}$}
        \STATE{$x_{\rm start} \gets x_{\rm i+1 \text{ mod } 100}$}
        \STATE{$y' \gets c_{i+1 \text{ mod } 100}$}                
        \ENDFOR
    \end{algorithmic}
\end{algorithm}


We use the classifiers defined in Table~\ref{tab:short} in our evaluation. 
All classifiers process input of the size $[0,1]^{3 \times 224 \times 224}$,
apart from Inception v3, which processes input of size $[0,1]^{3 \times 299 \times 299}$. 
Further, all classifiers expect input to be normalized with RGB color-channel means 
$(0.485, 0.456, 0.406)$ and standard deviations $(0.229, 0.224, 0.225)$, whereas
Inception v3 expects input to be normalized to range $[0,1]$ in all color channels. 
We re-normalize data to the correct range before processing it each classifier,
and ensure that images are of correct size with bilinear interpolation.

\begin{table}[t]
\caption{Shorthand notation for DNNs.}
\label{tab:short}
\begin{center}
\begin{small}
\begin{sc}
\begin{tabular}{clc}
\toprule
\textbf{} & \textbf{Classifier name (acronym)} & Input size \\
\midrule
1 & SqueezeNet 1.0 (SN1.0)~\cite{SqueezeNet} & 224 \\ 
2 & SqueezeNet 1.1 (SN1.1)~\cite{SqueezeNet} & 224 \\ 
3 & ResNet-18 (RN18)~\cite{he2016deep}  & 224 \\ 
4 & ResNet-34 (RN34)~\cite{he2016deep}  & 224 \\ 
5 & ResNet-50 (RN50)~\cite{he2016deep}  & 224 \\ 
6 & DenseNet-121 (DN121)~\cite{huang2017densely}  & 224 \\ 
7 & DenseNet-169 (DN169)~\cite{huang2017densely}  & 224 \\ 
8 & DenseNet-201 (DN201)~\cite{huang2017densely}  & 224 \\ 
9 & VGG11~\cite{Simonyan14c}  & 224 \\ 
10 & VGG16~\cite{Simonyan14c}  & 224 \\ 
11 & ResNet-101 (RN101)~\cite{he2016deep}  & 224 \\ 
12 & ResNet-152 (RN152)~\cite{he2016deep}  & 224 \\
13 & Inception v3 (IncV3)~\cite{szegedy2016rethinking} & 299 \\
\bottomrule
\end{tabular}
\end{sc}
\end{small}
\end{center}
\end{table}

We define our target APIs and substitute model ensembles in Table~\ref{tab:ensembles}.
For target APIs, we chose Inception v3, ResNet-101, ResNet-152 and VGG16 to experiment with different architecture choices having a low error rate in ImageNet examples\footnote{\url{https://pytorch.org/docs/stable/torchvision/models.html}}. The choice for ensemble components is also shown in Table~\ref{tab:ensembles}. We choose to divide the target models and ensemble models in this way to study the effect of \ourname between similar architectures (ResNet models, VGG models) and different architectures (Inception as target model). We also formulate target models as $\texttt{API}'_{\rm k=1}$, i.e the attacker can only obtain top-1 output from the target model.
The ensemble components were chosen with the 
principle of adding components of different architecture families 
 (SqueezeNet, ResNet, VGG, DenseNet) until the graphics card memory (GeForce 1060, 6GB) was filled. 
 While we  
 could have an even created a stronger adversary by loading additional 
 models in RAM, we found that this represented a reasonably strong and efficient adversary. 

\begin{table}[t]
\caption{Ensemble models and target API used in this work. Shorthand notation from Table~\ref{tab:short}. }
\label{tab:ensembles}
\begin{center}
\begin{small}
\begin{sc}
\begin{tabular}{cc}
\toprule
\textbf{Target} & \textbf{Ensemble components}  \\
\midrule
$ \text{IncV3} $ & Models 1--10 \\
$ \text{RN101} $ & Models 1--10 \\
$ \text{VGG16} $ & Models 1--9, 11 \\
$ \text{RN152} $ & Models 1--11 \\
\bottomrule
\end{tabular}
\end{sc}
\end{small}
\end{center}
\end{table}

We investigate the methods shown in Table~\ref{tab:methods} in this work.
Methods 1 (\baseline)~\cite{dong2018boosting} and 4 (\pines)~\cite{ilyas2018black}
 are state-of-the-art 
methods among black-box transferable evasion methods, 
and FDM attacks on partial information APIs, respectively. 
We investigate how well 
\baseline can reach eventual success, by continuing to query
\texttt{API} until a specified query budget is exceeded or success is encountered. 
Additionally, two variants are investigated. 
\ourname is a straightforward 
adaptation of \baseline 
for the setting where $x_{\rm start} \neq x_{\rm goal}$.
\ournamerand is a variant of \ourname, 
using random subsets of the ensemble. 
We set the query limit to 100,000 queries for \pines, and 
1,000 for methods \baseline, \ourname and \ournamerand 
after a preliminary investigation.
To ensure consistency, all methods are evaluated with PyTorch. 
We imported the code for \pines 
from the authors' TensorFlow repository\footnote{\url{https://github.com/labsix/limited-blackbox-attacks}} 
with default parameters. Methods \baseline, \ourname and \ournamerand 
use MIFGSM with the suggested momentum parameter $\mu = 1$ and are evaluated with the same ensembles. 
We evaluate adversarial examples with maximum perturbation $\norm{x_{\rm adv} - x_{\rm goal}}_\infty \le \epsilon = 0.05$, 
and step size $0.005$ (in \baseline, \ourname and \ournamerand)
, meaning that at least 10 MIFGSM steps are
 taking before querying the API for the first time. 
Methods \baseline, \ourname and \ournamerand 
start querying target APIs only after reaching this perturbation. 
To make results comparable, we assume no knowledge of the resizing operator on each
target API. Instead, black-box attacker $\mathcal{A}_{\rm Bbox}$ produces input of the size $[0,1]^{3 \times 224 \times 224}$
in all methods. 

\begin{table}[t]
\caption{Evasion methods. Methods 1--3 use substitute model ensembles from
Table~\ref{tab:ensembles}, while Method 4's formulation does not make use of substitute models. 
Method 1 starts adversarial example creation from the goal image $x_{\rm goal}$, while
Methods 2--4 have separate goal images and start images. }
\label{tab:methods}
\begin{center}
\begin{small}
\begin{sc}
\begin{tabular}{cllc}
\toprule
\textbf{} & \textbf{Method} & \textbf{Grad. est.} {$\mathbf{\hat{\texttt{G}}_{\rm dnn'}}$} & \textbf{Start class of} $x_{\rm start}$   \\
\midrule
1 & \baseline & $\text{MIFGSM}$~\cite{dong2018boosting}, full ens.  & $\texttt{API}_{\mathcal{L}}'(x_{\rm goal})$ \\
2 & \ourname & $\text{MIFGSM}$~\cite{dong2018boosting}, full ens. &  $y'$ \\ 
3 & \ournamerand & $\text{MIFGSM}$~\cite{dong2018boosting}, subset ens. & $y'$ \\ 
4 & \pines & NES~\cite{wierstra2008natural} &  $y'$ \\ 
\bottomrule
\end{tabular}
\end{sc}
\end{small}
\end{center}
\end{table}

\subsection{Evaluation criteria}
We next define criteria that we use to 
compare evasion methods. 



\paragraph{Success:} A boolean value denoting whether
a targeted 
adversarial example $x_{\rm adv}$ created by method $m_i$, for $\texttt{API}'$ 
such that $ y' \leftarrow \texttt{API}'(x_{\rm adv})$, 
using at most $B$ queries. 
\emph{Success rate} refers to how often success occurred in an experiment. 

\paragraph{Pareto-efficiency:}
given certain evasion 
setting \texttt{s}, a set of methods $\{m_1, \dots, m_L \}$ 
 and a criteria metric $q (x_{\rm start}, x_{\rm goal}, y', \texttt{API}', m_i) = q(\texttt{s}, m_i)$, 
 a method $m_i$ is said to be pareto-efficient for setting \texttt{s} if
 \begin{equation}
q (\texttt{s}, m_i) \le q (\texttt{s}, m_j)~\forall i,j \in {1, \dots, L}, i \neq j ,
\label{eq:pareto}
 \end{equation}
 
 i.e. $m_i$ produces the smallest criteria metric $q$ for setting \texttt{s}. 
Pareto-efficiency is a \emph{descriptive} property: given an experiment, we may
 calculate this statistic in hindsight. 
 
\paragraph{Dominance:}
a method $m_{i*}$ is said to dominate other methods in a range $q \in [a, b]$ if
 \begin{equation}
 \begin{split}
\text{E} ( q (\texttt{s}, m_{i*}) ) <& \text{E} ( q (\texttt{s}, m_j) )~\forall {i*},j \in {1, \dots, L}, {i*} \neq j \\
q* &= \min_i q (\texttt{s}, m_i)~\forall i\in \{1, \dots, L\} \\
q* &\in [a, b] ,
\end{split}
\label{eq:dom}
 \end{equation}

where $\text{E}(\cdot)$ is the expectation operator, i.e. we expect that $m_{i*}$ will 
produce the smallest performance metric, given that the pareto-efficient choice 
produces a metric in the range $[a,b]$. 
We consider the performance criteria \texttt{q*} in this paper.
\texttt{q*} is the minimum number of queries to \texttt{API}' it requires to produce an
adversarial example that is classified as class $y'$ on \texttt{API}' given setup $s$.
Dominance is a \emph{predictive} property: given previous tests, we may extrapolate
future performance. 


\subsection{Baseline evaluations and basic agility}
\label{sec:basic}

\begin{table}
  \begin{center}
    \caption{Effectiveness of baseline black-box evasion methods \baseline and \pines, and an \emph{agile} adversary \enspines. \emph{Success rate} and 
    \emph{average} 
    number of queries required for success. }
    \begin{tabular}{l c c c} 
\hline
& One query & \multicolumn{2}{r}{Up to 100,000 queries}  \\
            & \baseline &  \pines  & \enspines \\ \hline       
$\text{IncV3}$ & 12\% : 1 & 88\% : 44158 & 89\%: 40029 \\ 
$\text{RN101}$ & 47\%: 1 & 89\%: 32864  & 96\%: 18874 \\
$\text{VGG16}$ & 47\%: 1 & 94\%: 28875 & 94\%: 17433 \\
$\text{RN152}$ & 58\%: 1  & 91\%: 34689 & 95\%: 14754 \\
       \hline    
    \end{tabular}
    \label{table:eq_effectiveness}
  \end{center}
\end{table}

We first evaluate the baseline methods \baseline and \pines and 
how an agile adversary can simply increase efficiency and effectiveness. 
We show these results in Table~\ref{table:eq_effectiveness}.
The success rate of \baseline on the first try is shown in the leftmost column 
(up to 58\% on RN152). 
The success rate on IncV3 is only 12\%. 
We attribute this to the resize operator in 
$\text{IncV3}$, which resizes input from 
$(3 \times 224 \times 224)$ to $(3 \times 299 \times 299)$ 
(cf. Table~\ref{tab:short}). 
For example, Xie et al. \cite{xie2018mitigating} too found that resizing and cropping 
operations can act as a form of defense against adversarial examples.

On the other extreme, \pines reaches approximately 90\% success rate on all 
target APIs with up to 100,000 queries. \pines takes between 28,000 and 44,000
queries in average to succeed. 

We argue that the pareto-efficient choice for $\mathcal{A}_{\rm Bbox}$ is to 
switch from one
method to another when it becomes apparent that the first method will not succeed. 
Such an \emph{agile adversary} can combine the previous methods: after the first query, 
is done through \baseline , the remaining 99,999 queries can be done with \pines.
We call this simple agile method \enspines. 
We see in Table~\ref{table:eq_effectiveness} that 
the success rate for \enspines is between 0--7 percentage points higher
than with \pines only. In these cases, \baseline efficiently (1 query) finds an adversarial example that \pines fails at. 
This occurs especially on RN101 and RN152. We suspect this occurs due to similarity
of some of the ensemble components (Table~\ref{tab:ensembles}). 
\enspines can decrease the average queries between 42 -- 58 \% on RN101, VGG16 and 
RN152,
i.e. models where \baseline produces satisfactory transferability. 


\subsection{\ourname effectiveness}
\label{sec:effectiveness}

\begin{table}
  \begin{center}
    \caption{Effectiveness of black-box evasion methods, \emph{success rate} and \emph{median} number of queries required for success. 
    }
    \label{table:effectiveness}
    \begin{tabular}{l c c c c c } 
\hline
 & \multicolumn{4}{c}{Up to 1000 queries} \\
            & \baseline & \ourname & \ournamerand  & \pines \\ \hline       
$\text{IncV3}$  & 26\%: 2 & 69\%: 11 & 75\%: 14 & 0\%: -\\ 
$\text{RN101}$  & 83\%: 1 & 88\%: 8~ & 93\%: 12 & 0\%: -  \\
$\text{VGG16}$  & 82\%: 1 & 89\%: 10 & 90\%: 13 & 0\%: -  \\
$\text{RN152}$  & 84\%: 1 & 95\%: 8~ & 96\%: 11 & 0\%: -  \\
       \hline    
    \end{tabular}
  \end{center}
\end{table}

Next we evaluate variants of \ourname and compare them to the baseline methods. 
We show success statistics for each 
of these four methods in Table~\ref{table:effectiveness}, 
given our experimental setup and a query budget of 1000 queries.
Columns are arranged in the order of methods presented in Table~\ref{tab:methods}. 
Columns 1 -- 3 represent query use with substitute models, which we
advocate in this paper. 
If we continue querying for up to 1000 times 
\baseline reaches a high success rate on RN101, VGG16 and RN152 (between 82\% -- 84\%). 
However, \baseline success with 1000 queries on IncV3 is significantly lower: only 26\%. 
\ourname and \ournamerand reach 69\% and 75\%  
success rates respectively on IncV3, while limited to the
same query budget as \baseline. 
By comparison, \pines does not reach success since the query budget is too low. 

By comparing Tables~\ref{table:eq_effectiveness} and~\ref{table:effectiveness}, we see that 
the success rate of \ournamerand is in fact very similar as \pines on RN101, VGG16 and RN152,
while for most of the cases requiring \emph{3 orders of magnitude fewer} queries. 
We also see that 
in most cases, methods that enable higher success rate do this at the expense
 of a higher number of median queries. 
This motivates us to study the comparative effectiveness of different methods. 
Knowledge of such patterns can help in developing more efficient agile attacks
than the one presented in Section~\ref{sec:basic}. 



\subsection{Pareto-efficiency of methods}
\label{sec:efficiency}

\begingroup
\centering
\begin{figure*}[htbp]
\centering
\subfigure[Inception v3]{\includegraphics[width=0.3\textwidth]{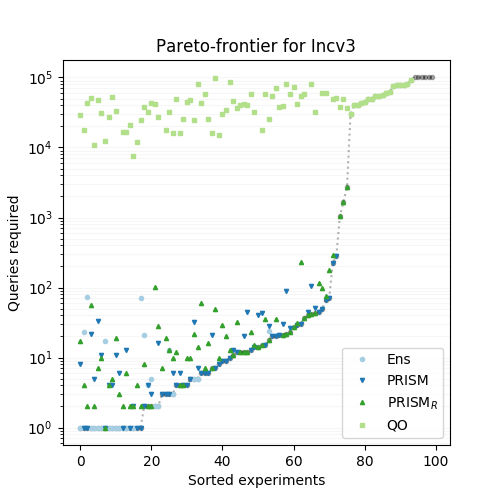}\label{fig:pareto-incv3}}
\subfigure[ResNet-101] {\includegraphics[width=0.3\textwidth]{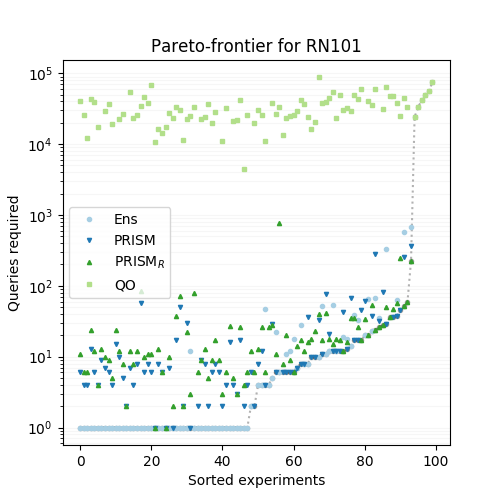}\label{fig:pareto-rn101}}
\subfigure[VGG16]{\includegraphics[width=0.3\textwidth]{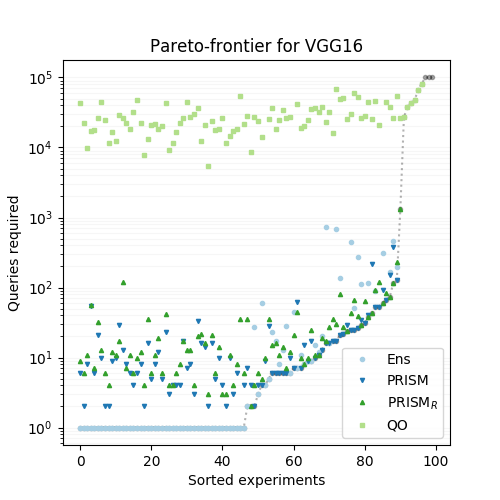}\label{fig:pareto-vgg16}}
\subfigure[Inception v3]{\includegraphics[width=0.3\textwidth]{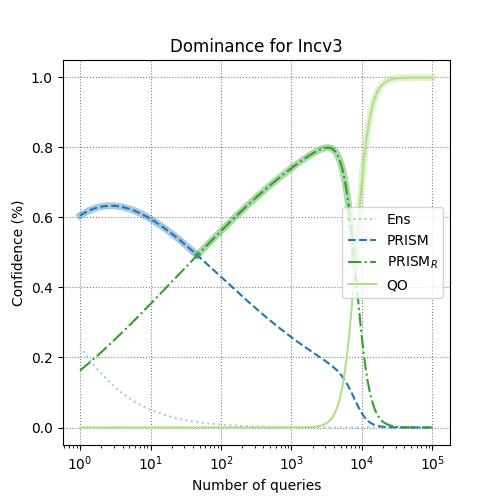}\label{fig:dominance-incv3}}
\subfigure[ResNet-101]{\includegraphics[width=0.3\textwidth]{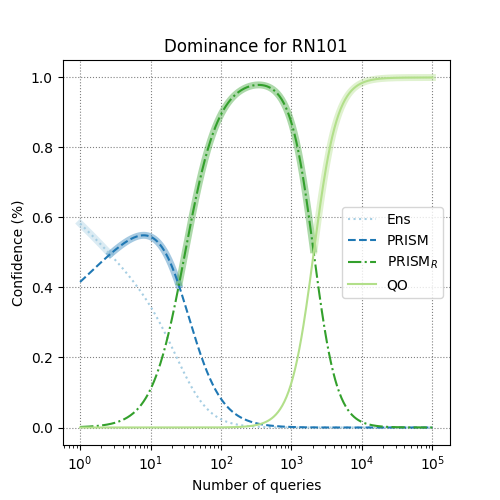}\label{fig:dominance-rn101}}
\subfigure[VGG16]{\includegraphics[width=0.3\textwidth]{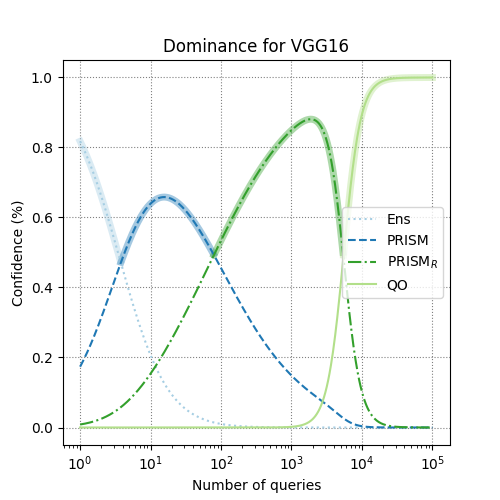}\label{fig:dominance-vgg16}}
\caption{Subfigures \ref{fig:pareto-incv3} -- \ref{fig:pareto-vgg16} illustrate \emph{pareto-efficiency} of \baseline (light blue dot), and \pines (light green square), \ourname (dark blue down-pointing triangle) and \ournamerand (dark green up-pointing triangle), evaluated against number of queries required to generate an adversarial example against three models: IncV3, RN101 and VGG16. 
Subfigures \ref{fig:dominance-incv3} -- \ref{fig:dominance-vgg16} illustrate dominance regions. Dominance is calculated as per Equation~\ref{eq:dom} and illustrates the most effective methods given a certain query region. 
The thick lines illustrate when methods are optimal and the confidence of that optimality. 
} \label{fig:pareto}
\end{figure*}

\endgroup

We show the query efficiency of the four methods in Figures~\ref{fig:pareto-incv3} 
to~\ref{fig:pareto-vgg16} on IncV3, RN101 and VGG16. 
The results are sorted so that the examples that required the least number of queries
are ordered to the left side. 
We see that some of the examples
are significantly harder than others (positive trend across colors). 
The most effective methods are connected
by a grey dotted line, denoting the pareto-efficient choices. 
We also see a progression that the most efficient methods
on the left side do not work efficiently on the right side.
It is \emph{harder} to find adversarial examples in some experiments than others:
by this we mean that the minimum number of 
queries required to create adversarial examples are bigger than in others.  
This hints at an inherent exploration-exploitation trade-off: methods that are the most 
efficient find the universally easy solutions quickly (good exploitation), but tend to underperform on more difficult tasks (poor exploration). 



\subsection{Dominance and efficient strategies}
\label{sec:strategy}

Factoring out the trivial case of transferability, we may ask what is the optimal 
evasion method, given a certain ``hardness'' of the task. 
Using the data points in the previous figure, we can predict which method performs 
 best given a certain region of required queries. 
Dominance (Equation~\ref{eq:dom}) can be the treated as a multiclass classification problem:
finding the most efficient evasion method, given a certain region
of required queries. We solve the problem with multinomial logistic regression.
Dominance regions are shown in Figures~\ref{fig:dominance-incv3} to~\ref{fig:dominance-vgg16}, 
and give hints about when it is sensible to switch evasion  
algorithms, calculated separately for each target model.

\begin{table}[h]
  \begin{center}
    \caption{Approximate dominance regions of evasion methods in Figures~\ref{fig:pareto-incv3} 
to~\ref{fig:pareto-vgg16} on IncV3, RN101 and VGG16. 
    }
    \label{table:dom}
    \begin{tabular}{c c c c } 
\hline
\baseline & \ourname & \ournamerand  & \pines \\ \hline       
0--1  & 1--50 & 50--1,000 & 1,000--100,000  \\
\hline    
    \end{tabular}
  \end{center}
\end{table}

We identify approximate dominance regions in Table~\ref{table:dom}. 
The results indicate a optimal progression of methods to try out, in the order of 1 to 4. 
Using this we can calculate several instantiations of 
\emph{efficient attacker strategies}, 
e.g. \EPPrQ tries \baseline for the first query, \ourname during queries  2--50, \ournamerand between 51--1000 and the rest with \pines. 
Following this strategy, we calculate that the average number of queries required to
create adversarial examples on each target model in Table~\ref{table:compare}. 
We compare this strategy to only using \EPPr and the previously presented 
\enspines and \pines. 

\begin{table}
  \begin{center}
    \caption{
    Comparison on effectiveness and efficiency of \EPPrQ compared to \EPQ , \enspines and \pines. 
    Column 1 details the \emph{success rate} and \emph{average} number of queries
    for success. 
     Columns 2 -- 4 compare this to alternative attacker strategies. 
     Row-wise best results are bolded.
    }
    \begin{tabular}{l c | c c c} 
\hline
& \EPPrQ & Cmp. \EPQ & Cmp. \enspines & Cmp. \pines \\
\hline 
$\text{IncV3}$ & \textbf{94}\%:{13477} & \textbf{$\pm$0 pp}: 1.12$\times$& -5 pp: 1.97$\times$ & -6 pp: 2.27$\times$\\
$\text{RN101}$ & \textbf{100}\%:{2882} & -1 pp: 1.56$\times$ & -5 pp: 6.55$\times$ & -11 pp:11.40$\times$ \\
$\text{VGG16}$ & \textbf{97}\%:~ {3497} & \textbf{$\pm$0 pp}: \textbf{1.00}$\times$& -3 pp: 4.98$\times$ & -3 pp: 8.26$\times$\\
$\text{RN152}$ & \textbf{100}\%:1419 & -1 pp: 1.36$\times$ & -5 pp:10.39$\times$ & -9 pp:24.43$\times$ \\

       \hline    
    \end{tabular}
    \label{table:compare}
  \end{center}
\end{table}

\EPPrQ is the most effective strategy. It reaches between \emph{94\% -- 100\% success rate}
on the evaluated models, which is betwen 3 and 11 points higher than \pines alone,
while using between 2.27$\times$ and 24.43$\times$ less queries. 
\EPPrQ is also more efficient than \enspines, while being significantly more effective. 
It is clear that \ourname is helpful towards increasing success rate and reducing
queries. 

We also compare \EPPrQ to \EPQ, to confirm whether \ournamerand has any impact
on \EPPrQ. We see that the inclusion of \ournamerand does not impact the 
success rate, but does impact the average number of queries on RN101 and RN152. 


\begin{table}
  \begin{center}
    \caption{
    Comparison on effectiveness and efficiency of \EPPr compared to \enspines and \pines . 
    Comparative success rate \emph{success rate} and 
    \emph{average} number of queries
    for success. 
     Row-wise best results are bolded.    
    }
    \begin{tabular}{l c | c c} 
\hline
& \EPPr & Cmp. \enspines & \pines \\ 
\hline 
$\text{IncV3}$ & 74\%: \textbf{107} & \textbf{+14 pp}: 387$\times$ & +13 pp: 427$\times$ \\ 
$\text{RN101}$ & 94\%:~ \textbf{30} & \textbf{+1 pp}: 667$\times$ & -5pp: 1162$\times$ \\ 
$\text{VGG16}$ & 91\%:~ \textbf{37} & \textbf{+3 pp}: 496$\times$ & \textbf{+3 pp}: 798$\times$ \\ 
$\text{RN152}$ & \textbf{98}\%: \textbf{28} & -3 pp : 530$\times$ & -7 pp: 1246$\times$\\ 
       \hline    
    \end{tabular}
    \label{table:compare2}
  \end{center}
\end{table}

We additionally compare an attack strategy that only relies on gradient estimates
from substitute model ensembles: \EPPr. 
We compare this strategy to \enspines and \pines in Table~\ref{table:compare2}. 
In one case out of four, \EPPr is most effective, and beats all strategies in
efficiency: 
it uses 2.6--2.8 orders of magnitude fewer queries than
the basic agile attack \enspines, and \emph{2.6--3.1 orders of magnitude fewer queries
 than \pines}. 

With these results we wish to highlight that agile attackers can present a realistic
threat to prediction APIs. 
We next discuss the threat that \ourname poses to real-life prediction APIs.

\subsection{Applicability to real-life APIs}
\label{sec:real}

As a proof-of-concept, we tested \ourname against Google Cloud Vision (GCV) API\footnote{Real-time attack demo: \url{https://www.dropbox.com/s/dzebo14v864sah2/Screencast\%202019-05-14\%2019\%3A05\%3A37.mp4?dl=0}}. 
GCV does not exactly fit our adversary model (Sect.~\ref{sec:adv-model}),
as it is trained with non-public data and uses different labels than the substitute models have.
This means that the approximation in Relation~\ref{eq:nabla} does not hold well. 
As we saw in Section~\ref{sec:efficiency}, some setups are harder than others. 
To provide a comparison with previous techniques, we took the same goal image as 
in Ilyas et al.~\cite{ilyas2018black}, with the task of changing the image classification of 
an image with two skiers to ``Canidae'' (dog-like mammal),
and set  $\epsilon=0.10$ (as in ~\cite{ilyas2018black}). 
However, GCV is under development and becomes more robust to adversarial examples. 
For example, the original attack example in Ilyas et al.~\cite{ilyas2018black} does not fool GCV anymore. 
For the the attack on GCV, we thus relaxed 
the optimization step S3 (Algorithm~\ref{alg:pi-algorithm}), and queried all 
intermediate crafted samples (thus requiring at a minimum 40 queries to go from $d=0.50$ to $d=\epsilon=0.10$ with step size $0.01$). 

\begin{figure}[htbp]
\centering
\includegraphics[width=0.45\textwidth]{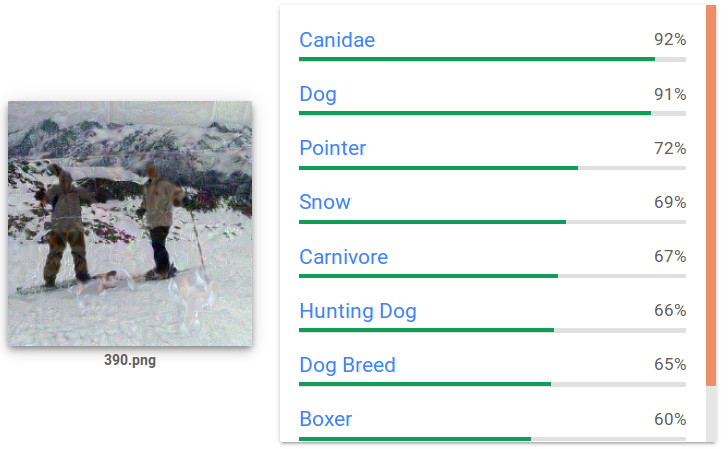}
\caption{
Adversarial example by \ournamerand against GCV (May 14th, 2019, $\epsilon=10\%$). Found
using 390 queries. 
} \label{fig:gcv}
\end{figure}

Nevertheless, we still succeeded. Figure~\ref{fig:gcv} shows the adversarial 
image that fools GCV in May 2019. The attack required in average 500 queries.
However, 
we found a large difference in the efficiency of our attack between January 2019 and
May 2019. In January 2019, we found that approximately 400 queries were enough to fool GCV with
modifications of size $\epsilon=5\%$. 

Even with 500 queries, the monetary cost of producing adversarial examples can 
be quite cheap, e.g. with current pricing approximately \$0.60 per example. This
is small compared to the previous approach by~\cite{ilyas2018black}, which 
required approximately \$26 with current pricing. 
We estimate that the cost of producing adversarial examples against real-world
APIs can be further dropped by judiciously optimizing the use of substitute model
ensembles and agile attacks. 

\section{Discussion}
\label{sec:discussion}
\label{sec:ensemblesize}

Next, we study the effect of the size of the ensemble and attack effectiveness. 
In order to understand the relationship between the number of ensemble components 
and the effectiveness of substitute model attacks like \ourname and \baseline, 
we calculated the success rate and the median  
number of queries required for success for different ensemble sizes in 
Table~\ref{tab:enscomp}. 
We increased the number of ensemble components step-by-step from one to ten. 
  We added models from Table~\ref{tab:short} to the ensemble in the order of  
 their ImageNet validation set accuracies\footnote{\url{https://pytorch.org/docs/stable/torchvision/models.html}}. 
 As usual, we evaluate adversarial examples with $\epsilon = 5\%$. 

We make the following observations from Table~\ref{tab:enscomp}. 
For a fixed $\epsilon$, 
%
we see 
that adding more components to the ensemble both \emph{increases the success rate} 
while \emph{decreasing} the median \emph{number of queries}, for all attacks. 
Adding components is helpful, even when the component itself might have fairly
 low accuracy (SN1.1 \& SN1.0). 
Table~\ref{tab:enscomp} also shows that the effectiveness of the attack increases rapidly when a component with a similar architecture is added to the ensemble (bold font). 
For example, when attacking VGG16, adding VGG11 to the ensemble increases the
 success rate from 54\% to 85\%. 
 We see that \ourname performs better than \baseline when several ensemble components are used. 
We provide intuition as to why \ourname performs better compared to \baseline in Appendix~\ref{sec:appendix}. 


\begin{table*}
\begin{center}
\caption{
Ablation study on the size of ensembles for \ourname and \baseline. 
The ensemble size is gradually increased from only one (left) to ten (right). 
Components with highest top-1 accuracy evaluated first.
Success rate and median number of queries for success shown. 
}
\begin{small}
\begin{tabular}{l c c c c c c c c c c}
Number of components. & 1 & 2 & 3 & 4 & 5 & 6 & 7 & 8 & 9 & 10 \\
\midrule
Target model & \multicolumn{10}{c}{IncV3} \\
\midrule
added model to ens. & DN201 & RN101 & RN50 & DN169 & DN121 & RN34 & RN18 & VGG11 & SN1.1 & SN1.0\\
\baseline (1 query)  & 2\% : 1 &  4\%: 1  & 5\%: 1   & 6\%: 1    & 6\%: 1   & 8\%: 1 & 10\%: 1 & 12\%: 1
  & 12\%: 1  & 12\%: 1 \\
\baseline (up to 1000 queries)   & 4\% : 9 &  6\%: 1  & 7\%: 1   & 10\%: 1    & 13\%: 2   & 14\%: 1 & 18\%: 1 & 24\%: 1
  & 23\%: 1  & 26\%: 2 \\
\ourname (up to 1000 queries) & 2\%: 89 & 6\%: 60 & 12\%: 28 & 16\%: 27 & 26\%: 14 & 41\%: 15 & 54\%: 12 & 60\% 10 & 62\%: 9 & 69\%: 11 \\
\midrule
Target model & \multicolumn{10}{c} {RN101} \\
\midrule
added model to ens. & DN201 & \textbf{RN50} & DN169 & DN121 & \textbf{RN34} & VGG16 & \textbf{RN18} & VGG11 & SN1.1 & SN1.0\\
\baseline (1 query)   & 4\%: 1 &  11\%: 1 & 14\%: 1  & 21\%: 1  & 34\%: 1 & 34\%: 1  & 39\%: 1 & 45\%: 1 & 44\%: 1  & 47\%: 1 \\
\baseline (up to 1000 queries)   & 7\%: 1 &  \textbf{29\%}: 8 & 35\%: 3  & 43\%: 2  & 59\%: 1 & 62\%: 1  & 74\%: 1 & 76\%: 1 & 78\%: 1  & 83\%: 1 \\
\ourname (up to 1000 queries) & 7\%: 84 & \textbf{34\%}: 29 & 49\%: 26 & 56\%: 16 & \textbf{69}\%: 14 & 69\%: 12 & \textbf{83}\%: 8 & 
85\%: 8 & 87\%: 9 & 88\%: 8 \\
\midrule

Target model & \multicolumn{10}{c}{VGG16} \\
\midrule
added model to ens. & DN201 & RN101 & RN50 & DN169 & DN121 & RN34 & RN18 & \textbf{VGG11} & SN1.1 & SN1.0\\
\baseline (1 query)  & 1\%: 1 &  3\%: 1 & 3\%: 1  & 6\%: 1  & 13\%: 1  & 13\%: 1  & 21\%: 1 & \textbf{41}\%: 1  & 42\%: 1  & 47\%: 1  \\
\baseline (up to 1000 queries)   & 6\%: 11 &  9\%: 87 & 13\%: 17  & 20\%: 10  & 26\%: 1  & 34\%: 5 & 44\%: 2 & \textbf{75}\%: 1  & 80\%: 1   & 82\%: 1 \\
\ourname (up to 1000 queries) & 3\%: 248 & 7\%: 77 & 15\%: 52 & 27\%: 34 & 32\%: 22 & 38\%: 21 & 54\%: 17 & 
\textbf{85}\%: 12 & 86\%: 11 & 86\%: 10 \\
\midrule
\end{tabular}
\end{small}
\label{tab:enscomp}
 \end{center}
\end{table*}

\section{Related work}
\label{sec:related}
Our paper explored a limited knowledge \emph{substitute learner} adversary model~\cite{munoz2017towards}.
Other adversary models have also been considered:

\paragraph{Limited knowledge \emph{surrogate data}:}

Tramer et al~\cite{tramer2016stealing}, Papernot et al~\cite{papernot2017practical}, Li et al.~\cite{pengcheng2018query} and Juuti et al~\cite{juuti2018prada} develop model extraction attacks against DNNs by training a substitute model using synthetic data. Labels are obtained by querying the target model. 
The substitute model is later used to form transferable adversarial examples.

\paragraph{Model confidentiality:}
There have been several attempts~\cite{gilad2016cryptonets, liu2017oblivious, juvekar2018gazelle} 
to make the DNNs APIs \emph{oblivious}, such that the API can process inputs correctly
without learning anything about the client's input, the client simultaneously does
not learning anything about the model behind the API.

\paragraph{Other black-box attacks using substitute models:}
Wang et al~\cite{wang2018great} explore transfer learning, where a student model for a specific application initialized by a publicly available pretrained teacher model (e.g., Inception v3, VGG16 etc). They show that one can compute adversarial perturbations that can mimic hidden layer representations copied from the teacher in order to fool the student model. They also used ensembles in the case of student model knowledge and an unknown teacher model. 
They show that their attack performance degrades if several layers of the student
models are fine-tuned. 
Ji et al~\cite{ji2018model} maliciously train pre-trained models in order to implement model-reuse attacks against ML systems 
without knowing the developer's dataset or fine-tuning strategies.
Hashemi et al~\cite{hashemi2018stochastic} query the target model with images that come from a similar distribution as the training images of the target model, augment the dataset with random noise and use this augmented dataset to train a substitute model.  They craft adversarial examples against the substitute model using Carlini \& Wagner~\cite{carlini2017towards} method and
perform transferability attacks. 
However, for training substitute models, they require logits from the target model in order to mimic decision boundaries and this attack can fail in case of more limited information.
Guo et al~\cite{guo2018countering} implemented non-differentiable image transformation techniques as a preprocessor in order to defend against black-box and gray-box model evasion attacks. This type of gradient obfuscation techniques are effective when the adversary do not have the knowledge about the preprocessing method. 

\paragraph{Finite-difference method attacks:}
Similarly to us, Du et al~\cite{du2018towards} also consider the partial information attack,
and separate between start and goal images. However, they initialize the starting 
image with a gray color and adopt NES for gradient estimation. 
They attacked a cloud API (Clarifai food detection) by choosing a valid label 
from top-k classes and minimizing the probability of so-called non-object or 
 background predictions. Although their gray-image attack requires fewer queries 
 than typical finite-difference methods as in ~\cite{chen2017zoo, ilyas2018black}, 
 the adversarial examples are unrecognizable by humans, which is different from our case.
Brendel et al~\cite{Brendel2017a} introduce a decision-based attack 
which initializes the starting sample that is already adversarial and walks along the boundary between
the adversarial and non-adversarial 
region as well as decreasing the distance towards the target image. 
They only used top label for initializing the starting image and 
finding the direction along the boundary, which is similar to
our evaluations, 
but their attack requires 
more than an order of magnitude
more iterations than the attacks evaluated in this work. 
Other publications have used gradient-free optimization techniques, 
such as genetic algorithms~\cite{alzantot2018genattack} or 
greedy local search~\cite{narodytska2017simple} 
over the image space in order to craft adversarial image in a black-box setting.

\section{Conclusions}
\label{sec:conclusions}
We presented targeted evasion attacks using substitute model ensembles for black-box
APIs. 
We showed that such attacks can achieve very high effectiveness and efficiency:
reaching similar effectiveness as state-of-the-art finite-difference attacks
on partial-information APIs, while requiring up to 3 orders of magnitude fewer queries.
We showed that the attack relies on the appropriateness of an implicit gradient estimation,
and that this gradient approximation benefits from larger substitute model ensembles. 
Query use with ensembles seems like an interesting direction to explore for future research. 
As a proof-of-concept, the partial-information formulation of the attack was used to attack Google Cloud Vision, where it succeeded in approximately 400 queries. 
We argue that query-limited substitute model attacks form a pervasive threat against present-day cloud APIs due to the availability of substitute models and relatively cheap pricing. 

\begin{acks}
This work was supported in part by the  Intel (ICRI-CARS). 
We thank Samuel Marchal and Sebastian Szyller for interesting
discussions, and Aalto Science-IT project for the computational resources.
\end{acks}


%
\bibliographystyle{ACM-Reference-Format}
\bibliography{ref}
\appendix
\section{Appendix}
\label{sec:appendix}

\begin{figure}[htbp]
\centering
\includegraphics[width=0.45\textwidth]{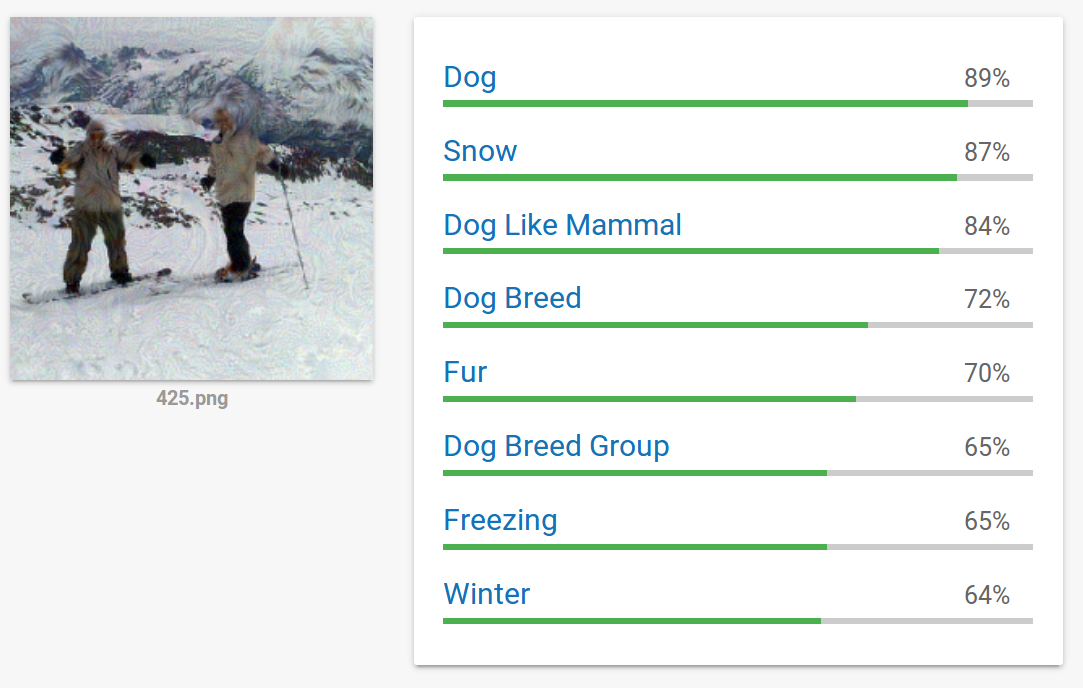}\label{fig:pert-baseline}
\caption{
Adversarial example by \ourname against GCV (January 7th, 2019, $\epsilon=5\%$).
Found using 425 queries. 
} \label{fig:gcv_old}
\end{figure}

Figure~\ref{fig:gcv_old} shows an evasion example created by \ourname in January, 2019.

\begingroup
\centering
\begin{figure*}[h]
\centering
\subfigure[$x_{\rm goal}$]
{\includegraphics[width=0.25\textwidth]{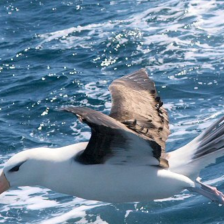}\label{fig:pert-goal}}
\subfigure[$x_{\rm adv}$ found by \ourname]{\includegraphics[width=0.25\textwidth]{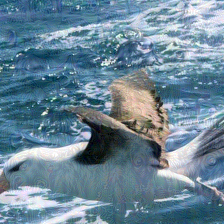}\label{fig:pert-adv}}
\subfigure[$x_{\rm start}$]{\includegraphics[width=0.25\textwidth]{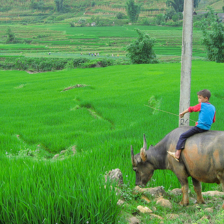}\label{fig:pert-start}}
\caption{
An example of an adversarial example $x_{\rm adv}$ found by \ourname, given goal image $x_{\rm goal}$ and start image $x_{\rm start}$. Created against against IncV3 with perturbation $\epsilon = 5\%$.  
}  \label{fig:adv_example}
\end{figure*}
\endgroup

\begin{figure*}[h]
\centering
\subfigure[(Closed) linear span of $(x_{\rm start}, x_{\rm adv})$ found by \ourname. Confined to range ((-0.1, 0.5), (0. 0.1)). \ourname finds an $x_{\rm adv}$ where a linear interpolation maintains the same classification region as $x_{\rm start}$.]{\includegraphics[width=.8\textwidth]{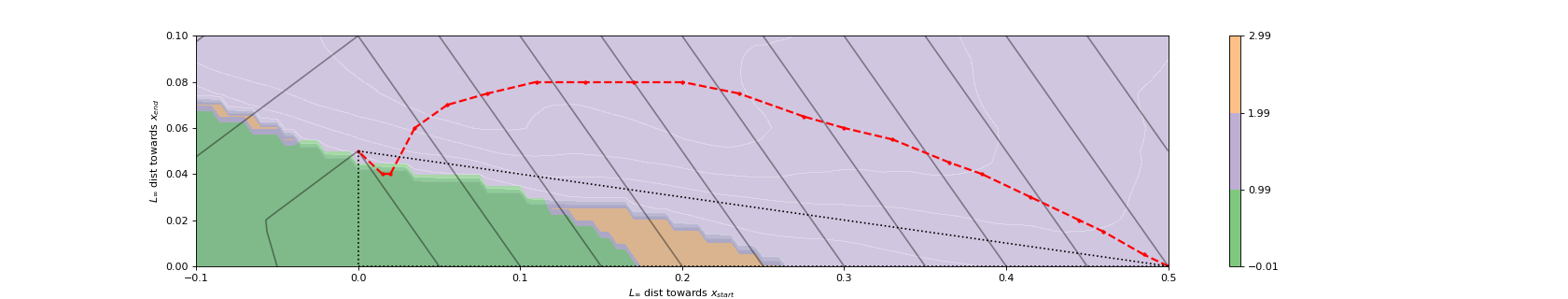}\label{fig:topology-rn101-piens}}
\caption{
Example of linear spans of $(x_{\rm start}, x_{\rm adv})$, starting with the same input, comparing two methods. The sample creation starts in the lower right corner (0.5,0) and progresses towards (0,0). The process ends when the coordinate (0, 0.05) is reached and the sample is still in the same classification region as in $x_{\rm start}$. 
Evaluated with RN101. Green marks the original class (sea gull), purple the target class (water buffalo) and orange other classes.  Contour regions inside the purple region mark logit values of RN101. Red dashed lines mark the path from $x_{\rm start}$ to $x_{\rm adv}$, projected down to the closest point in the linear span of $(x_{\rm start}, x_{\rm adv})$, in terms of $L_1$-distance. 
Diagonal lines marks successive 5\% absolute increments in $L_\infty$-distance: from $x_{\rm start}$. Classification regions sampled with resolution 121x21.
} \label{fig:topology}
\end{figure*}

\begingroup
\centering
\begin{figure*}[h]
\centering
\subfigure[\baseline]{\includegraphics[width=0.25\textwidth]{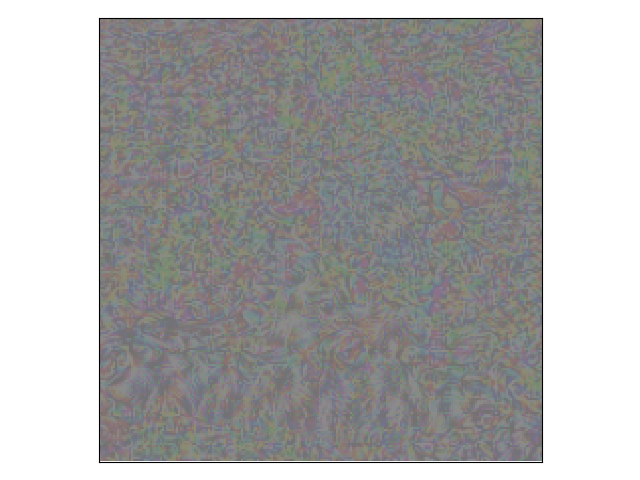}\label{fig:pert-baseline}}
\subfigure[\pines]{\includegraphics[width=0.25\textwidth]{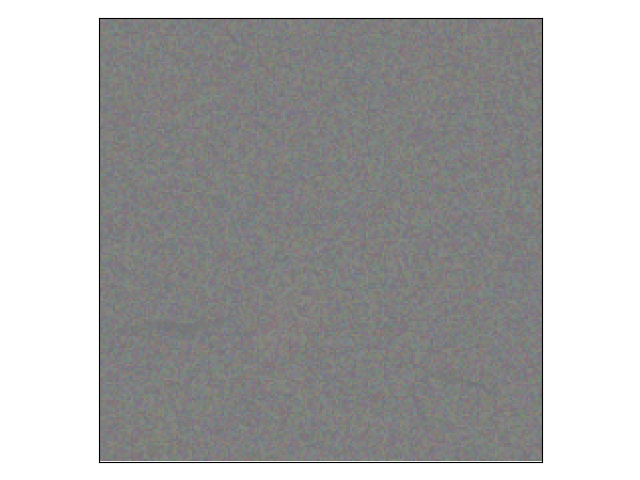}\label{fig:pert-pines}}
\subfigure[\ourname]{\includegraphics[width=0.25\textwidth]{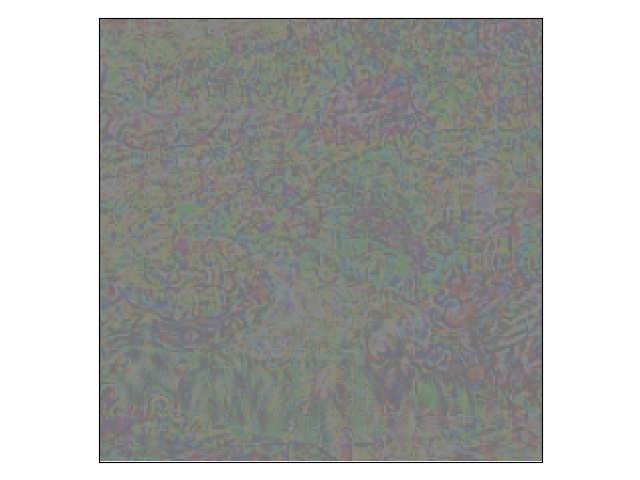}\label{fig:pert-piens}}
\caption{
Perturbations of adversarial examples. Created against against IncV3 with perturbation $\epsilon = 5\%$.  
} \label{fig:pert}
\end{figure*}
\endgroup

We show an example of how \ourname evades black-box APIs next.
 Figure~\ref{fig:adv_example} shows a resulting adversarial example, given a goal
 image and a start image. Figure~\ref{fig:topology} shows the process of finding
 the adversarial example with \ourname. 
 The path that \ourname induces is shown in Figure~\ref{fig:topology-rn101-piens}. 
 Although it is a black-box attack, it essentially follows a hill-climbing 
 route due to the similarity of the gradients of the substitute model and target model. 
The results in Figure~\ref{fig:topology-rn101-piens} suggest that a path between $x_{\rm start}$ and $x_{\rm adv}$ may be found without breaking the classification region of DNNs. 
The success behind 
 \ourname implies that DNNs have connected but complex classification regions. These results reinforce empirical results by Fawzi et al~\cite{fawzi2017classification} who claim that classification regions of DNNs  form connected regions, rather than isolated pockets.




We compare perturbations created by \baseline, \pines and \ourname in Figure~\ref{fig:pert},
evaluated with IncV3. The goal image and start image are the same as in Figure~\ref{fig:pert-goal} and~\ref{fig:pert-start}. 
\pines perturbations resemble random noise, 
whereas perturbations created via \baseline  
and \ourname contain regular grid-resembling structures. 
The perturbation found by \ourname additionally 
contains localized perturbations influenced from $x_{\rm start}$ (Figure~\ref{fig:pert-start}),
as suggested by Figure~\ref{fig:topology-rn101-piens}. 
Note the gradient at the gull wing, greenish tint in background and retained buffalo head.



\end{document}